\documentclass[runningheads]{llncs}
\usepackage{graphicx}
\usepackage[linesnumbered]{algorithm2e}
\usepackage{lscape} % for rotating table
\usepackage{float} % for [p] at table
\usepackage{graphicx} % for making table smaller
\usepackage{comment}
\usepackage{afterpage}
\usepackage{textcomp} % for arrows
\usepackage[utf8]{inputenc} \usepackage[T1]{fontenc} % for i with trema
\usepackage{lmodern}  % for i with trema
\usepackage[USenglish]{babel}
\usepackage{censor}
\usepackage{rotating}
\PassOptionsToPackage{hyphens}{url}\usepackage{hyperref}

\usepackage{fourier} 
\usepackage{array}
\usepackage{makecell}

\begin{document}

\title{RDF2Vec Light -- A Lightweight Approach\\for Knowledge Graph Embeddings\thanks{Copyright \textcopyright~2020 for this paper by its authors. Use permitted under Creative Commons License Attribution 4.0 International (CC BY 4.0).}}
\titlerunning{RDF2Vec Light}

\author{Jan Philipp Portisch\inst{1,2}\orcidID{0000-0001-5420-0663} \and
Michael Hladik\inst{2}\orcidID{000-0002-2204-3138} \and
Heiko Paulheim\inst{1}\orcidID{0000-0003-4386-8195}}

\authorrunning{J. P. Portisch et al.}

\institute{Data and Web Science Group, University of Mannheim, Germany\\
	\email{\{jan, heiko\}@informatik.uni-mannheim.de} \and
	SAP SE Product Engineering Financial Services, Walldorf, Germany\\
\email{\{jan.portisch, michael.hladik\}@sap.com}}

\maketitle
\setcounter{footnote}{0}
\pagenumbering{gobble}
\begin{abstract}
Knowledge graph embedding approaches represent nodes and edges of graphs as mathematical vectors. Current approaches focus on embedding complete knowledge graphs, i.e. all nodes and edges. This leads to very high computational requirements on large graphs such as \textit{DBpedia} or \emph{Wikidata}. However, for most downstream application scenarios, only a small subset of concepts is of actual interest. In this paper, we present \textit{RDF2Vec Light}, a light\-weight embedding approach based on \textit{RDF2Vec} which generates vectors for only a subset of entities. To that end, \textit{RDF2Vec Light} only traverses and processes a subgraph of the knowledge graph. Our method allows the application of embeddings of very large knowledge graphs in scenarios where such embeddings were not possible before due to a significantly lower runtime and significantly reduced hardware requirements. 

\end{abstract}
\section{Introduction}
Public knowledge graphs (KGs), such as \emph{DBpedia} or \emph{Wikidata}, provide deep background knowledge that can be exploited for downstream tasks such as question-answering or recommender systems~\cite{heist2020knowledge}. \textit{KG embeddings} (KGEs) represent vertices and, depending on the approach, also edges of a KG as numeric vectors. This representation is easily consumable by most algorithms and can be exploited in downstream tasks. Advantages of KGEs, once they have been trained, include simple applicability, fast run time, good performance on multiple tasks, and reusability in downstream applications. On the downside, KGEs produce very large models\footnote{For example, the 200 dimensional DBpedia RDF2Vec embedding model available at \textit{KGvec2go}~\cite{kgvec2go} requires more than 10GB of disk storage.}, and are very expensive to train and re-train in the case of evolving knowledge bases. For very large knowledge graphs, such as Wikidata, computing a complete embedding typically takes up to a day or longer~\cite{han2018openke}.

In this paper, we address the scalability aspect of knowledge graph embeddings: Our novel approach, \emph{RDF2Vec Light}, allows to train partial, task-specific models with only a fraction of the computation requirements compared to other embedding approaches, while retaining a high performance on multiple tasks. The resulting models contain \emph{only vectors for entities of interest}. Internally, \emph{RDF2Vec Light} only traverses a subset of the underlying knowledge graph which leads to processing times that are much shorter than the original RDF2Vec approach which always processes an entire knowledge graph.
Moreover, the resulting models are much smaller.\footnote{\emph{RDF2Vec Light} models are typically only a few kilobytes in size, compared to multiple gigabytes of disk space required to persist classic embedding models.} 

\begin{algorithm}[t]
 \KwData{\textit{G = (V,E)}: RDF Graph, $V_I$: vertices of interest, \textit{d}: walk depth, $n$: number of walks}
 \KwResult{ $W_G$: Set of walks }
 $W_G = \emptyset$\\
 \For{vertex v $\in V_I$}{
    \For{ 1 to $n$}{
    
     add $v$ to $w$\\
     $pred$ = getIngoingEdges(v)\\
     $succ$ = getOutgoingEdges(v)\\
     \While{$w$.length() $< d$}{
            $cand$ = $pred$ $\cup$ $succ$\\
            $elem$ = pickRandomElementFrom($cand$)\\
            \If{$elem$ $\in$ $pred$}{
                add $elem$ at the beginning of $w$\\
                $pred$ = getIngoingEdges($elem$)
            }
            \Else{
                add $elem$ at the end of $w$\\
                $succ$ = getOutgoingEdges($elem$)
            }
        }
        add $w$ to $W_G$\\
    }
 }
 \caption{Walk generation algorithm for \textit{RDF2Vec Light}.}
 \label{alg:walk_generation}
\end{algorithm}
 
\section{RDF2Vec Light}
\label{sec:approach}
\emph{RDF2Vec} is based on performing random walks on a graph~\cite{rdf2vec_journal}. The underlying idea of \textit{RDF2Vec Light} embeddings is to generate only local walks for entities of interest given a predefined task. After the walk generation has been completed, the training of vectors can be performed like in the original approach. 

Rather than \emph{starting} random walks at all entities of interest, it is randomly decided for each depth-iteration whether to go backwards, i.e. to one of the node's predecessors, or forwards, i.e. to the node's successors (line 9 of Algorithm~\ref{alg:walk_generation}). As a result, the entities of interest can be at the beginning, at the end, or in the middle of a walk which better captures the context of the entity. This generation process is described in Algorithm~\ref{alg:walk_generation}.
The RDF2Vec method as well as the \textit{RDF2Vec Light} extension have been implemented in Java and Python.\footnote{\url{https://github.com/dwslab/jRDF2Vec}} The implementation can handle various RDF formats such as n-triples, RDF/XML, Turtle, or HDT~\cite{hdt_reference}. In addition, a REST API has been implemented and is provided on \url{http://www.kgvec2go.org}.

\section{Evaluation}
\label{sec:evaluation}
\begin{table}[t]
    \caption{Classification (accuracy) and regression (RMSE) results with RDF2Vec Classic and \emph{RDF2Vec Light}. The best classic and light results are highlighted.}
    \label{tab:results}
    \centering
    \begin{tabular}{l|r|r|r|r|r|r|r|r|r|r}
	&	\multicolumn{2}{c|}{Cities}	&	\multicolumn{2}{c|}{Movies}	&	\multicolumn{2}{c|}{Albums} &	\multicolumn{2}{c|}{AAUP} &	\multicolumn{2}{c}{Forbes} \\
Strategy	&	SVM	&	LR	&	SVM	&	LR	&	SVM	&	LR	&	SVM	&	LR	&	SVM	&	LR	\\
\hline
Light\_500\_4\_CBOW\_50	&	52.56	&	19.23	&	73.31	&	19.75	&	72.44	&	12.52	&	61.93	&	68.35	&	60.79	&	34.62	\\
Classic\_500\_4\_CBOW\_50	&	49.36	&	16.95	&	55.25	&	22.77	&	51.70	&	14.06	&	55.33	&	70.59	&	57.28	&	36.64	\\
\hline
Light\_500\_4\_CBOW\_100	&	71.78	&	21.16	&	73.50	&	19.90	&	72.43	&	12.35	&	63.21	&	\textbf{65.85}	&	60.81	&	34.96	\\
Classic\_500\_4\_CBOW\_100	&	49.36	&	22.15	&	58.21	&	22.94	&	57.44	&	14.17	&	55.00	&	73.33	&	57.36	&	42.32	\\
\hline
Light\_500\_4\_CBOW\_200	&	71.61	&	54.86	&	73.93	&	19.60	&	73.34	&	12.50	&	61.74	&	67.65	&	59.11	&	35.97	\\
Classic\_500\_4\_CBOW\_200	&	49.36	&	99.73	&	58.79	&	23.54	&	59.18	&	14.24	&	56.83	&	80.29	&	57.57	&	45.76	\\
\hline
Light\_500\_4\_SG\_50	&	\textbf{75.90}	&	\textbf{19.39}	&	74.15	&	19.34	&	76.49	&	12.00	&	\textbf{65.46}	&	67.66	&	\textbf{61.56}	&	34.58	\\
Classic\_500\_4\_SG\_50	&	\textbf{\underline{80.57}}	&	\textbf{\underline{12.95}}	&	72.81	&	19.89	&	76.42	&	11.80	&	\textbf{\underline{68.04}}	&	64.85	&	61.08	&	\textbf{34.89}	\\
\hline
Light\_500\_4\_SG\_100	&	73.99	&	20.89	&	\textbf{\underline{74.89}}	&	\textbf{\underline{19.21}}	&	\textbf{\underline{76.98}}	&	\textbf{11.89}	&	64.54	&	66.59	&	61.38	&	\textbf{\underline{34.48}}	\\
Classic\_500\_4\_SG\_100	&	79.01	&	15.26	&	72.72	&	\textbf{19.61}	&	\textbf{76.51}	&	\textbf{\underline{11.57}}	&	64.72	&	\textbf{\underline{65.50}}	&	60.42	&	35.26	\\
\hline
Light\_500\_4\_SG\_200	&	73.81	&	44.38	&	74.58	&	19.45	&	76.35	&	12.16	&	62.83	&	70.13	&	60.26	&	36.73	\\
Classic\_500\_4\_SG\_200	&	77.06	&	28.34	&	\textbf{73.85}	&	19.71	&	75.66	&	11.92	&	66.74	&	67.96	&	\textbf{\underline{61.82}}	&	36.93	\\
    \end{tabular}
\end{table}
In order to evaluate the approach presented in this paper, the classification and regression experiments, as well as the entity and document relatedness experiments of Ristoski et al.~\cite{rdf2vec_journal} have been repeated. The evaluation follows the setup defined in~\cite{DBLP:conf/esws/PellegrinoCGR19}.

Six classic and six light embedding spaces have been trained each with the following parameters held constant: $window$ $size$ $= 5$, $negative$ $samples $ $= 25$. The parameters that were changed are the generation mode ($cbow$ and $sg$) as well as the dimension of the embedding space ($50$, $100$, $200$). All walks have been generated with 500 walks per entity and a depth of 4. For the evaluation, the DBpedia knowledge graph as of 2016-10\footnote{\url{https://wiki.dbpedia.org/downloads-2016-10}} has been used. 

In the results tables, \emph{strategy} refers to the configuration with which the embeddings have been obtained. The structure can be read as follows:\\ \textit{<mode>\_}\textit{<number\_of\_walks\_per\_entity>\_}\textit{<walk\_}\textit{depth>\_}\textit{<training\_mode>\_<dimen\-sion>} where \textit{mode} is either \textit{Light} or \textit{Classic}.

For the classification and regression tasks, we follow the same setup as in the original RDF2Vec paper~\cite{DBLP:conf/semweb/RistoskiVP16}: For the classification tasks, four classifiers have been evaluated: \emph{Na\"{i}ve Bayes}, \emph{C4.5} (decision tree algorithm), \emph{k-NN} with $k=3$, and \emph{Support Vector Machines (SVM)} with $C \in \{10^{-3}, 10^{-2}, 0.1, 1, 10, 10^{2}, 10^{3}\}$ where the best $C$ is chosen. A 10-fold cross validation has been used to calculate the performance statistics.
For the regression tasks, three approaches have been evaluated: \emph{linear regression}, \emph{k-NN}, and \emph{M5rules}. For the sake of brevity, we only report results for the best performing approaches (SVM and LR).\footnote{The complete result tables are available at \url{http://www.rdf2vec.org/rdf2vec_light}}

\begin{table}[t]
\caption{Results on the document relatedness task (LP50), reporting the harmonic mean of Pearson correlation and Spearman rank correlation, and on entity relatedness (KORE), using cosine similarity. The best value of each comparison group is highlighted in bold. The overall best value is additionally underlined.}
\label{tab:document_entity_relatedness}\centering
\begin{tabular}{l|r|r||l|r|r}
Strategy & LP50 & KORE & Strategy & LP50 & KORE\\
\hline
Light\_500\_4\_CBOW\_50 & 0.3871 & 0.3343 & Light\_500\_4\_SG\_50 & 0.3421 & 0.4767\\
Classic\_500\_4\_CBOW\_50 & 0.2235 & 0.2982 & Classic\_500\_4\_SG\_50 & 0.4400 & 0.5068\\
\hline
Light\_500\_4\_CBOW\_100 & 0.3741 & 0.3179 & Light\_500\_4\_SG\_100 & 0.3310 & 0.4281\\
Classic\_500\_4\_CBOW\_100 & 0.2291 & 0.3028 & Classic\_500\_4\_SG\_100 & 0.4507 & 0.5288\\
\hline
Light\_500\_4\_CBOW\_200 & 0.3809 & 0.3474 & Light\_500\_4\_SG\_200 & 0.3278 & 0.4045\\
Classic\_500\_4\_CBOW\_200 & 0.2374 & 0.3178 & Classic\_500\_4\_SG\_200 & 0.4371 & 0.5348\\
\end{tabular}
\end{table}

For classification and regression, we can observe that except for the cities dataset, the difference between the two approaches is rather marginal. For entity and document relatedness, the results are less conclusive. Here, we see that the RDF2Vec light approach is en par with the classic approach for the CBOW variant, but the results are reversed when looking at the SG variant, which also yields the best results globally.

In order to distinguish the cases where RDF2Vec Light is en par with its classic counterpart from those where it is clearly inferior, we looked at the linkage degree of the entities at hand, as well the homogeneity of the entities of interest. 

We can observe that a higher degree of the entities of interest leads to a worse performance of \emph{RDF2Vec Light}, as in the inferior performance of \emph{RDF2Vec Light} for the cities datasets in classification and regression: Cities are among the most strongly interlinked entities in DBpedia~\cite{heist2020knowledge}. Likewise, the document and entity similarity datasets contain a larger number of strongly interlinked head entities.

\begin{figure}[t]
\centering
\includegraphics[width=\textwidth]{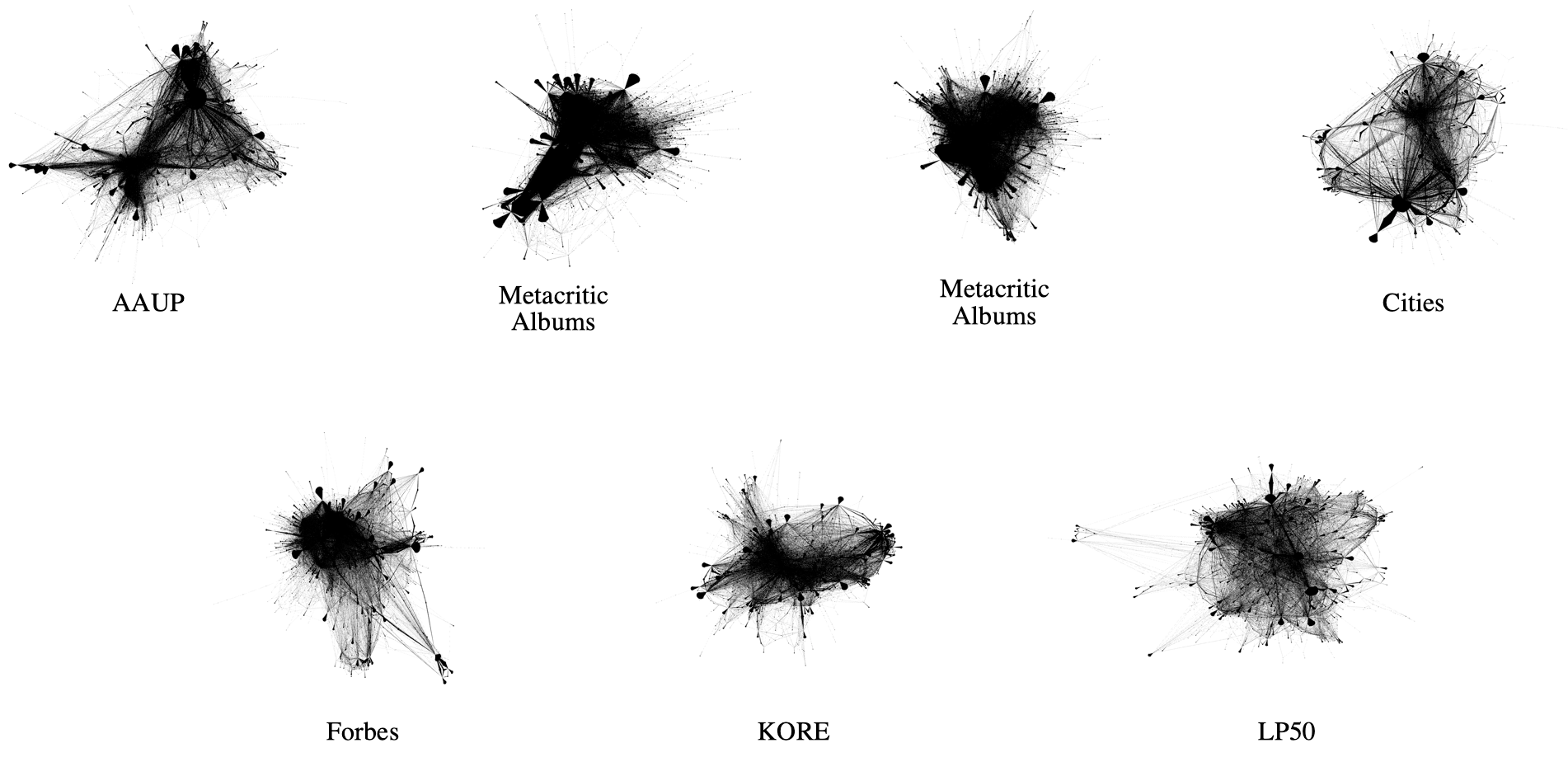}
\caption{Depiction of the graphs which were assembled using the generated walks.}
\label{fig:assembled-graphs-rendered}
\end{figure}

While for classification and regression problems, the set of entities is rather homogeneous (i.e., all are cities, albums, etc.), the homogeneity is lower for the document and entity relatedness, where the entities of interest are scattered across many classes. Both degree and homogeneity contribute to the density of the considered subgraphs, as depicted in Fig.~\ref{fig:assembled-graphs-rendered}. From the plots, we can observe a correlation of \emph{RDF2Vec Light} performance and the density of the graph spanned by the random walks -- the more dense the graph (i.e., the less head entities there are and the more homogeneous the entity set at hand), the better the performance of \emph{RDF2Vec Light}.

The runtime of \emph{RDF2Vec Light} is linear in the number of entities of interest. On commodity hardware, the runtime is roughly 1 minute per 10 nodes. In comparison, training RDF2Vec on the full DBpedia graph takes a few days.

\section{Conclusion and Outlook}
In this paper, we presented \textit{RDF2Vec Light}, an approach for learning latent representations of knowledge graph entities that requires only a fraction of the computing power compared to other embedding approaches. Rather than embedding the whole knowledge graph, \textit{RDF2Vec Light} trains vectors for only few entities of interest and their context. For this approach, the walk generation algorithm has been adapted to better represent the context of the entities. Our experiments show that the results achieved with \emph{RDF2Vec Light} are comparable to those obtained with the standard RDF2Vec, while requiring only a fraction of the runtime.

\bibliographystyle{splncs04}
\bibliography{references}

\end{document}